# BREAST ANATOMY ENRICHED TUMOR SALIENCY ESTIMATION

*Fei Xu[1], Yingtao Zhang[2], Min Xian[3], H. D. Cheng[1,2], Boyu Zhang[1],
Jianrui Ding[4], Chunping Ning[5], Ying Wang[6]*

[1] Department of Computer Science, Utah State University, Logan, USA
[2] School of Computer Science and Technology, Harbin Institute of Technology, Harbin, China
[3] Department of Computer Science, University of Idaho, Idaho Falls, USA
[4] School of Computer Science and Technology, Harbin Institute of Technology, Weihai, China
[5] Department of Ultrasound, Affiliated Hospital of Medical College Qingdao University, Qingdao, China
[6] Department of General Surgery, Second Hospital of Hebei Medical University, Shijiazhuang, China

## ABSTRACT

Breast cancer investigation is of great significance, and developing tumor detection methodologies is a critical need. However, it is a challenging task for breast ultrasound due to the complicated breast structure and poor quality of the images. In this paper, we propose a novel tumor saliency estimation model guided by enriched breast anatomy knowledge to localize the tumor. Firstly, the breast anatomy layers are generated by a deep neural network. Then we refine the layers by integrating a non-semantic breast anatomy model to solve the problems of incomplete mammary layers. Meanwhile, a new background map generation method weighted by the semantic probability and spatial distance is proposed to improve the performance. The experiment demonstrates that the proposed method with the new background map outperforms four state-of-the-art TSE models with increasing 10% of $F_{meansure}$ on the BUS public dataset.

## 1. INTRODUCTION

More than 2 million women are diagnosed with breast cancer every year, and more than 620,000 will die from the disease [1]. The early detection and treatment of breast cancer will increase the survival rate greatly [2,3].

In clinical routine, breast ultrasound (BUS) is a primary modality for cancer screening [4, 5], and automatic BUS image segmentation methods are essential for cancer diagnosis and treatment planning. Many automatic BUS segmentation approaches have been studied [5-9]. However, the performances of the models were instable due to collected images under various sources and periods using different machines with various qualities of the images, such as low contrast, more artifacts, etc. [8-9] proposed BUS segmentation models based on deep neural networks and [9] demonstrated that the CNN models could achieve much better performance than the traditional models. However, the two challenges existed: 1) no enough BUS image data available for training; 2) segmentation results completely based on the training dataset and the deep network.

VSE measures the probabilities of image regions attracting human attention, which is essential and accessible for detecting the objects and achieving automatic segmentation [10-15]. Current VSE approaches can be classified into two categories: bottom-up (data-driven) [10-12] and top-down (task-driven) [13-15] models. The former models use low-level features or prior knowledge, and they cannot handle complicated scenes because the low-level features or prior knowledge cannot present the high-level or semantic context in the images properly. Three strategies are employed in the most of CNN-based modes: 1) utilize more than one deep neural network to generate the saliency maps [14-15]; 2) integrate the high-level semantic knowledge by the deep neural network and low-level hand-craft features or visual saliency hypothesis [13,15]; 3) refine the object boundary in the final step [13]. The studies showed that CNN-based models generated much better performance than the bottom-up models.

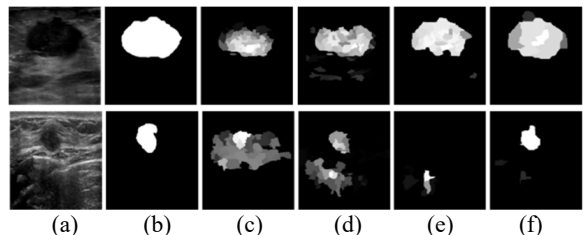

Fig. 1. Tumor saliency detection examples. (a) original images (b) Ground truths (c)-(f) the saliency maps generated by [6], [16], [17] and the proposed method, respectively.

For BUS images, many VSE methods have been investigated [6,16-18]. Examples are shown in Fig. 1. [6] applied the global contrast visual saliency hypothesis and tumor appearance prior to generate the saliency map. The method could locate the tumor accurately in most cases of its dataset. However, it missed parts of the big tumors and made the surrounding regions of tumors have high saliency values (see Fig.1(c)). [16] proposed an optimization framework to estimate the tumor saliency after determining the tumor existence and generated good performance on the public dataset. However, the optimization framework also made the non-tumor regions have high intensities as [6]. [17] presented a novel unsupervised framework to estimate the tumor saliency based on integrating breast anatomy modeling. It decomposed the BUS image into several horizontal layers by Neutro-Connectedness (NC) theory, which would make the regions with strong connectedness gather into the same layer. However, the generated layers can not present semantic anatomy information. In this paper, we propose a novel TSE top-down model. Firstly, we utilize U-Net [8] to generate the initial four semantic breast anatomy layers (skin, fat, mammary, and muscle layers) [9]. Then it refines the wrong breast anatomy layers by combining the non-semantic decomposing layers based on NC theory (refer [17]). The final saliency maps are generated by the optimization framework integrating foreground cue, background cue, adaptive-center bias, and region-based correlation.

The rest of the paper is organized as: section 2 describes the proposed approach; the experiments are explained in section 3; the conclusion and future work are discussed in section 4.

## 2. THE PROPOSED APPROACH

The proposed approach generates the tumor saliency map by the existing united optimization-based framework [17] integrating robust cognitive hypotheses, e.g., the adaptive center-bias, and region-based correlation, and the background and foreground cues. The saliency map is $S = (s_1, s_2, \cdots, s_N)^T$ which is a vector of saliency values, and $s_i$ denotes the saliency value of the $i$th region and $s_i \in [0, 1]$. $N$ is the number of superpixels generated by [19]. The optimization formulation is:

$$minimize\ E(S) = S^T\big(-(\alpha \ln(C) + \beta \ln(F))\big) +$$
$$\gamma(1-S)^T(-\ln(T)) + \sum_{i=1}^{N}\sum_{j=1}^{N}(s_i - s_j)^2 r_{ij} D_{ij} \quad (1)$$
$$\text{subject to } 0 \le s_i \le 1, i = 1, 2, \cdots, N;$$
$$B^T S = 0, B = (b_1, b_2, \cdots b_N)^T, b_i = \{0,1\}$$

$$r_{ij} = \exp(-|I'_i - I'_j|/\sigma_1^2) \quad (2)$$
$$D_{ij} = \exp(-\|rc_i - rc_j\|_2/\sigma_2^2) \quad (3)$$

In Eq. (1), the term $T = (t_1, t_2, \cdots, t_N)^T$ is the background map, and larger $t_i$ indicates the $i$th region belonging to the background with higher probability; the term $C = (c_1, c_2, \cdots, c_N)^T$ defines the coordinate distances between the regions' centers and the adaptive-centers, and larger $c_i$ indicates that the region is closer to the adaptive-center; the term $F = (f_1, f_2, \cdots, f_N)^T$ is the foreground map, and larger $f_i$ indicates the higher probability of the $i$th region belonging to the foreground, and the terms $r_{ij}$ and $D_{ij}$ define the similarity and the spatial distance between the $i$th and the $j$th regions, respectively. The term $(1-S)^T(-\ln(T))$ defines the cost on the background map and forces the regions with smaller values in the background map to have higher values in the saliency map. The term $S^T(-\ln(C))$ defines the cost of the adaptive-center bias and forces the regions with larger distances to have smaller values in the saliency map. The term $S^T(-\ln(F))$ defines the cost of the foreground map and forces the regions with smaller values in the foreground map to have smaller values in the saliency map. The quadratic term models the region-based correlations which force similar regions with similar saliency values. Parameters $\alpha$, $\beta$, and $\gamma$ are used to balance the impact of each component. In Eqs. (2) and (3), $|\cdot|$ is the $l_1$ norm, $\|\cdot\|_2$ is the $l_2$ norm, and $\sigma_1^2 = \sigma_2^2 = 0.5$ by [17].

### 2.1. Breast anatomy layers(BAL) generation

#### 2.1.1. Initial breast anatomy layers

The breast contains four primary layers: skin layer, fat layer, mammary layer, and muscle layer [9]. Regions in different layers have different appearances, and the tumor always exists in the mammary layer. Due to the limitation of the number of training data, it is a challenge to generate accurate tumor segmentation results based on CNN.

The proposed approach utilizes the well-known U-Net [8], which consists of fully convolutional encoder and decoder sub-networks with skip connections. [9] demonstrated that the U-Net could generate good performance on limited BUS images dataset for producing the initial SBAM. The number of convolutional filters in the network is (32, 32, 64, 64, 128). The input images have dimensions of 256×256 pixels, noted as $I$. The segmentation result of U-Net, $SA$, has dimension 256×256, and the segmentation probability map, $SP$, has dimension 4×256×256. $SP_{k,i,j}$ denotes the pixel in $SP$ and indicates the probability of pixel $I(i,j)$ belonging to the $k$th category. The value $SA_{i,j}$ is $k = \max(SP_{k,i,j}|_{k=0}^{3})$. The pixel-based maps $I$, $SP$ and $SA$ are converted into the region-based maps $I'$, $SP'$ and $SA'$ using the region-based optimization framework. The label with the largest value of the labels of each superpixel will be the region label. $SA'_i$ indicates the skin, fat, mammary, and muscle layer, and $i$ is 1, 2, 3, 4, respectively. More details will be discussed in section 3.1.

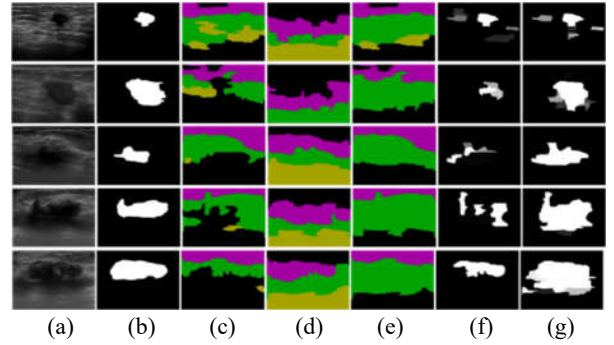

(a)    (b)    (c)    (d)    (e)    (f)    (g)

Fig. 2 The visual effects of refining BAL. (a) original images; (b) ground truths; (c) BAL generated by U-Net; (d) the non-semantic layers generated by [17]; (e) the refined BAL; (f) the FG based on (c); and (g) the FG based on (e).

#### 2.1.2. Refine breast anatomy layers

Based on observation, U-Net generates acceptable breast anatomy layers in most cases even using small training dataset. However, some anatomical layers missed most of the parts, and cross-layer ( one layer is divided into more than one part with no connections by other layers) appears in some cases. It refines the initial breast anatomy layers by the NC anatomical map, $NCL$ [17] which decomposes the BUS image into 3-5 horizontal layers by NC and the regions in the same layer with strong connectedness. The refined breast anatomy layers noted as $NSA$, contains the same 4 layers as $SA'$. $NCL_i$, $NSA_i$ and $SA'_i$ indicates the $i$th layer of $NCL$, $NSA$ and $SA'$, respectively. $NSA_1$ is the intersection between the $SA'_1$ and the union of first and last layer of $NCL$. If $SA'_2$ is valid (the layer covered more than 75% columns of the image [17]), find the layer $i$ of $NCL$ that contains the regions in $SA'_2$; use the regions in the layers $NCL_k$ where $k$ is less than $i$ and the union with $SA'_2$ excluding the regions in the other layers of $SA'$ as $NSA_2$; otherwise $NSA_2$ is different set between the rest of $NCL_1$ and the other layers of $SA'$. If $SA'_4$ is valid, find the layer $i$ of $NCL$ that contains the regions in $SA'_4$; use the regions in the layers $NCL_k$ where $k$ is greater than $i$ and the union with $SA'_4$ excluding the regions in the other layers of $SA'$ as $NSA_4$; otherwise, $NSA_4$ is the different set of the last layer of $NCL$ and $SA'_1$. The rest regions will be assigned to the mammary layer of $NSA_3$.

After refining, the $NSA$ maps are kept the same as $SA$ in most of the cases (see the 1st row of Fig. 2). The refinement is to avoid cross-layer and incomplete layer and to keep the high recall ratio on the mammary layer (see Fig. 2). It will reduce the probability of missing the tumor (see 2nd-5th rows of Fig. 2(c)) and recover the incomplete layer generated by the deep learning models (see Fig.2(e)).

### 2.2. Foreground map (FG) generation

The foreground map (FG) measures image regions' possibilities to be tumor regions. [17] proposed two algorithms to identify the dark/shadow layers and generate a foreground map for each layer, and it produced good results; especially, on the images with large and/or small tumors. Algorithm 1 in [17] outputs a flag with three values. If *flag*=-1, it indicates a smooth layer (most of the regions in the layers with high intensities); If *flag*=1, it indicates a dark/shadow layer; otherwise, it is a normal layer. We adopt the same algorithm to identify the dark layer and employ the Z-function to generate the FG.

### 2.3. Adaptive-center distance map generation

[14] proposed the adaptive-center bias instead of the fixed image center bias, which estimated the adaptive center (AC) using a weighted local contrast map on natural images. [16, 17] demonstrated the effectiveness of generating the AC by weighted foreground map on BUS images. In this paper, we adopt the method to generate the AC and the distance map $C$. $c_i = \exp(-\|rc_i - AC\|_2/\sigma_3^2)$ where the $rc_i$ is the center coordinate of the *i*th region; $\|\cdot\|_2$ is the $l_2$ norm; and $\sigma_3^2 = 0.1$ [17].

### 2.4. Background map (BG)

Boundary connectivity is effective prior to many visual saliency estimation models [11-13,16,17]. [16,17] have demonstrated the boundary connectivity based on NC theory, which calculated NC between the regions and the boundary regions to avoid noisy data and to generate much smoother and more accurate background map on BUS images [20]. Therefore, we generate the NC map by the algorithm in [20] and denote $nc_i$ as the NC value of the *i*th region in the NC map. The value of the *i*th region in the initial BG map is defined as $t_i = nc_i^2$.

Meanwhile, it defines the layer weight according to the region-based semantic probability maps $SP'$ (refer 2.2.1). The initial weight for each layer is defined as the mean value of the probabilities of the regions in the layer belonging to the mammary layer (Eq. (4)). If the mammary layer is valid (the layer covered more than 75% columns of the image [17]), it assigns $rW_{k,i} = \max(SP'_{3,i}, LayerW_k)$; otherwise, assigns all the regions with 1.

$$LayerW_k = \frac{\sum_{i \in NSA_k} SP'_{3,i}}{|NSA_k|} \quad (4)$$

where $LayerW_k$ is the probability of the *k*th layer belonging to the mammary layer; $SP'_{3,i}$ is the predicted probability of the *i*th region belonging to the mammary layer; and $rW_{k,i}$ is the probability of the *i*th region in the *k*th layer belonging to the mammary layer.

To avoid the isolated region in the non-mammary layer having very small *t* value, it defines the final value of the *i*th region in the initial BG map weighted by the probability of mammary layer and the distance from the AC, as $t_i = 1 - (1 - nc_i^2) \times rW_{k,i} \times c'_i$. $c'_i$ is 1 if $flag = 1$ in the mammary layer and $c_i > 0.5$; or $flag = 0$ in the non-mammary layers and $c_i \geq 0.75$; otherwise $c'_i = c_i$. T is normalized. The newly defined background map avoids the situation that some isolating non-tumor regions obtain the lowest values and decrease the saliency values of the tumor regions too much even the tumor regions gain the highest values in the foreground map (see Fig. 3). The effectiveness of the new background map will be discussed in section 3.3.

### 2.5. Optimization

The optimization framework is similar to that in [16,17], and it utilizes the same optimization method with the same initial and stops conditions in [17].

Fig. 3 shows the final optimal saliency maps generated with different components in the objective function. The model with the BG in [17] will decrease the tumor saliency values when non-tumor regions gain the lowest value in BG (see Fig.3(f)). The overall performance will be discussed in section 3.3.

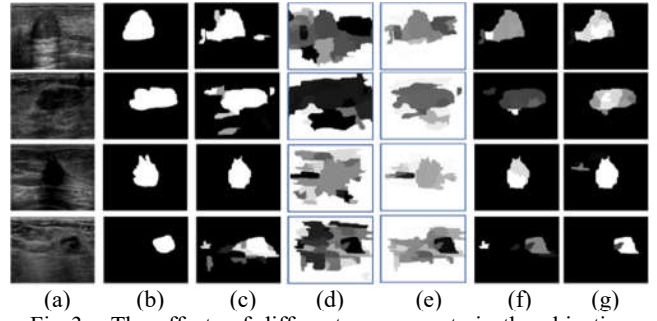

(a)  (b)  (c)  (d)  (e)  (f)  (g)

Fig 3  The effects of different components in the objective function. (a) original images; (b) ground truths ; (c) the FG; (d) the BG map in [17]; (e) the proposed BG; (f) the saliency map based on (d); (g) the saliency map using new BG.

## 3. EXPERIMENTS

### 3.1 Datasets, metrics and setting

Using a dataset of 325 images, 229 images contain tumors, and the other 96 images have no tumors [9]. The training and validation subsets are randomly chosen: 90% images from the total dataset (the images with tumor and without tumor are 90%, respectively), and the ratio of the training set and validation is 8:2. The rest 10% dataset is used as the test subset. The training epochs are 100, the batch size is 5, and learning rate is 0.1.

It validates the performance of the newly proposed TSE method using a dataset containing 562 BUS images from a public benchmark [21]. For tuning the parameters in Eq. (1), it randomly chooses 60 images as a training set, and the rest is utilized to evaluate the overall performance.

**Metrics of saliency estimation**: It uses Precision-recall (P-R) curve, mean Precision and Recall rate, $F_{meansure}$ and mean absolute error (*MAE*) to evaluate the performance. For each method, it normalizes the intensities of the saliency map into [0, 255]. Then it banalizes the saliency map by the threshold ranging from 0 to 255 and computes the precision and recall rates by comparing the thresholding result with the ground truth. The P-R curve is calculated by averaging precision-recall ratios of the dataset. The precision and recall ratios are defined as follows:

$$Precison = \frac{|SM \cap GT|}{|SM|}, Recall = \frac{|SM \cap GT|}{|GT|}$$

where *SM* is the binary saliency map, *GT* is the ground truth, and $|\cdot|$ denotes the number of pixels of values 1s. To obtain the average precision and recall ratios, it uses an adaptive thresholding method [22], which chooses two times the mean saliency value as the threshold. The $F_{meansure}$ [10] and *MAE* [22] are defined as

$$F_{meansure} = \frac{(1 + \theta^2)Precison \cdot Recall}{\theta^2 \cdot Precison + Recall}$$

$$MAE = \sum_{i=1}^{P}|SM(p_i) - GT(p_i)|$$

where $\theta^2$ is set to 0.3 as in [10], $p_i$ is the coordinate of the *i*th pixel, $SM(p_i)$ is the saliency value of the *i*th pixel, and *GT* is the binary ground truth. The value of each pixel in *SM* or *GT* is between 0 to 1. A good algorithm will obtain a smaller *MAE* and a larger $F_{meansure}$.

**Parameter setting:** all the experiments are based on $\boldsymbol{\alpha} = 10, \boldsymbol{\beta} = 51$, and $\boldsymbol{\gamma} = 6$.

### 3.2 Parameters tuning

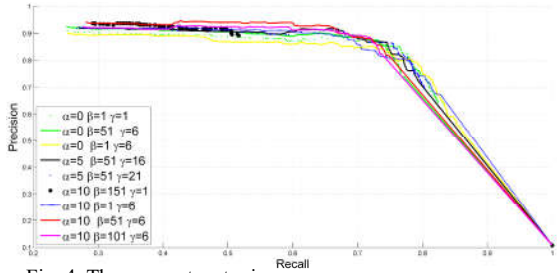

Fig. 4. The parameters tuning.

As presented in section 2, there are four components in the objective function. The parameters $\alpha$, $\beta$, and $\gamma$ balance the impact of each component and generate better performance. We evaluate the performance of the proposed method using randomly selected subset of 60 images from the 562 images [21] using different parameters and choosing each parameter which could obtain the better P-R curve and *MAE* value when the P-R curve is similar. Since the objective function is similar to that in [17], it adopts the initial ranges for the three parameters in [17]; and the performances are similar when the parameter on the foreground map is less than 50, and the parameter on the background map is less than 10. Therefore, the range $\alpha$ is from 0 to 10 with step size 5, $\beta$ from 1 to 151 with step size 50 and $\gamma$ from 1 to 21 with step size 5, respectively. As shown in Fig. 4, the P-R curves are competitive under most of the parameter combinations, and it achieves a better P-R curve and *MAE* when $\alpha$ is 10, $\beta$ is 51, and $\gamma$ is 6, respectively.

### 3.3 The overall performance of the proposed method

The proposed model is compared with most recently published TSE methods SMTD [6], HFTSE [16], TBAM [17], and two models generated by the proposed method with the background maps generated by different strategies on the 502 images, SMTD, HFTSE, and TBAM are the bottom-up VSE models with the specific breast tumor appearance knowledge. SMTD defined a unified global contrast mapping to estimate the tumor saliency. HFTSE proposed an optimization TSE model after determining the existence of a tumor. The proposed method is denoted as OURs, with the newly proposed BG map. OUR_BG1 is the optimization model with the BG map generated by HFTSE in Eq. (1) and the FG map with local contrast strategy.

The comparison visual effects of detecting saliency map by the five models are shown in Fig. 5. OUR_BG1 obtains a similar saliency map in most of the cases that the tumor regions gain the lowest values in the background map. However, it will only highlight the non-salient regions and decrease the saliency values of the tumor regions when there are some isolating non-tumor regions having the lowest values in the background map (see Fig. 3 and the 1st -3rd rows of Fig. 5). SMTD would miss some parts of big objects (see the 1st and 2nd rows of Fig. 5) and make the surround dark regions have high saliency values (see the 3rd -6th rows of Fig. 5). This situation will make the model SMTD achieve a high recall ratio but low precision ratio (see Fig.6). HFTSE would miss parts of large tumors and make the surrounding tumor regions have high saliency values (see the 2nd -6th rows of Fig. 5). TBAM model generated accurate saliency maps when detecting the correct layer having tumors (see the 1st, 2nd and 4th rows of Fig. 5 ); especially, for the images with large or small tumors; however, it failed in the cases that the tumors are in the top or bottom of the images (see the 2nd row of Fig.1 and 5th row of Fig. 5 ).

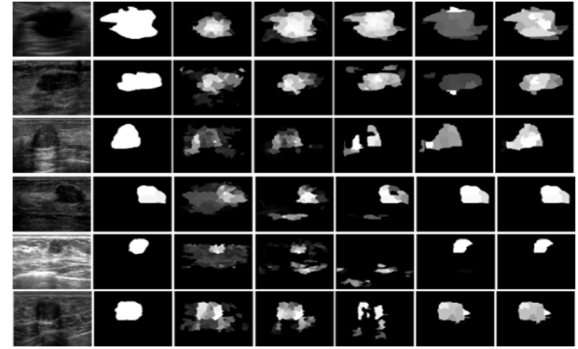

(a)   (b)   (c)   (d)   (e)   (f)   (g)

Fig. 5. The visual effects of detecting the saliency maps by the five models. (a) original images;(b)ground truths; (c)-(g) the saliency maps generated by [6], [16], [17], OUR_BG1 and OURs, respectively.

The overall performances of the seven models are shown in Figs. 6. The proposed model, OURs, achieves the best P-R curve, lowest *MAE*, and highest $F_{meansure}$ values. OUR_BG1 generates the competitive P-R curve as OURs. However, OURs achieves much better $F_{meansure}$ values, which indicates that a better background map generation will improve the TSE performance a lot.

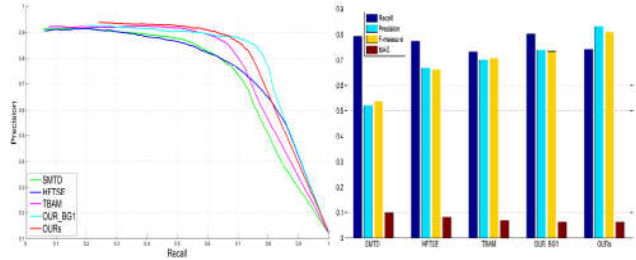

Fig. 6. The P-R curves, *MAE* and $F_{meansure}$ values of the five models.

### 4. CONCLUSION

In this paper, we propose a novel TSE model by utilizing the semantic breast anatomy knowledge. In the model, non-semantic breast anatomy modeling is integrated to solve the cross-layer and incomplete mammary layer in the SBAM. The strategy is effective even when the semantic information could not be generated accurately due to limited data. A new background map generation method is proposed to improve the performance, which is weighted by the semantic probability and spatial distance in the mammary layer. The experiment demonstrates that the background map has a great impact on the performance, and the model with the new background map improves the overall performance a lot. The proposed method outperforms four state-of-the-art TSE models on the datasets. In the future, we will focus on generalizing the proposed semantic-based model to other image modalities and diseases.

### 5 ACKNOWLEDGMENTS

This work was partially supported by the Center for Modeling Complex Interactions (CMCI) at the University of Idaho through NIH Award #P20GM104420.